%% file: main.tex
\documentclass[10pt,twocolumn,letterpaper]{article}

\usepackage{cvpr}              
\input{preamble}
\definecolor{cvprblue}{rgb}{0.21,0.49,0.74}
\usepackage[pagebackref,breaklinks,colorlinks,allcolors=cvprblue]{hyperref}
\usepackage{multirow}
\usepackage[accsupp]{axessibility}  
\newcommand{\cwh}[1]{{\color{black} { #1}}} 

\makeatletter
\def\@fnsymbol#1{}
\makeatother

\def\paperID{43567} 
\def\confName{CVPR}
\def\confYear{2026}

\title{Parameter-Efficient Semantic Augmentation for \\ Enhancing Open-Vocabulary Object Detection}

\author{
Weihao Cao\textsuperscript{1,2,3,\textdagger}\hspace{0.5em}
Runqi Wang\textsuperscript{2,3,\textdagger}\hspace{0.5em}
Xiaoyue Duan\textsuperscript{4,\textdagger}\hspace{0.5em}
Jinchao Zhang\textsuperscript{4}\hspace{0.5em}
Ang Yang\textsuperscript{1,2,3}\hspace{0.5em}
Liping Jing\textsuperscript{1,2,3,*}\thanks{\textdagger\ Equal contribution. * Corresponding author.} \\
\textsuperscript{1}State Key Laboratory of Advanced Rail Autonomous Operation, Beijing Jiaotong University \\
\textsuperscript{2}Beijing Key Laboratory of Traffic Data Mining and Embodied Intelligence, Beijing Jiaotong University \\
\textsuperscript{3}School of Computer Science \& Technology, Beijing Jiaotong University \quad
\textsuperscript{4}WeChat AI, China \\
}

\vspace{-10mm}
\begin{document}
\maketitle
\input{sec/0_abstract}    
\input{sec/1_intro}
\input{sec/2_related}
\input{sec/3_method}
\input{sec/4_experiments}
{
    \small
    \bibliographystyle{ieeenat_fullname}
    \bibliography{main}
}

\end{document}

%% file: sec/0_abstract.tex
\begin{abstract}
\vspace{-7mm}

\cwh{
Open-vocabulary object detection (OVOD) enables models to detect any object category, including unseen ones. Benefiting from large-scale pre-training, existing OVOD methods achieve strong detection performance on general scenarios (e.g., OV-COCO) but suffer severe performance drops when transferred to downstream tasks with substantial domain shifts.
This degradation stems from the scarcity and weak semantics of category labels in domain-specific task, as well as the inability of existing models to capture auxiliary semantics beyond coarse-grained category label.
To address these issues, we propose HSA-DINO, a parameter-efficient semantic augmentation framework for enhancing open-vocabulary object detection. 
Specifically, we propose a multi-scale prompt bank that leverages image feature pyramids to capture hierarchical semantics and select domain-specific local semantic prompts, progressively enriching textual representations from coarse to fine-grained levels.
Furthermore, we introduce a semantic-aware router that dynamically selects the appropriate semantic augmentation strategy during inference, thereby preventing parameter updates from degrading the generalization ability of the pre-trained OVOD model.
We evaluate HSA-DINO on OV-COCO, several vertical domain datasets, and modified benchmark settings.
The results show that HSA-DINO performs favorably against previous state-of-the-art methods, achieving a superior trade-off between domain adaptability and open-vocabulary generalization. The code is available at \href{https://github.com/jokercao6/HSA-DINO}{github.com/jokercao6/HSA-DINO}.
}

\end{abstract}

%% file: sec/1_intro.tex
\vspace{-5mm}
\section{Introduction}

\begin{figure}[h]
    \centering
    \includegraphics[width=0.9\columnwidth]{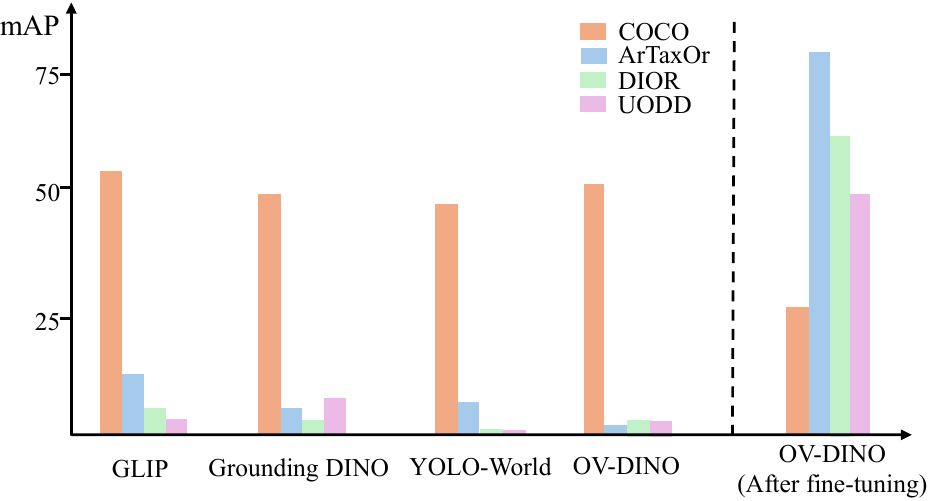} 
    \vspace{-2.5mm}
    \caption{
    Pre-trained OVOD models perform well on general domain (\emph{e.g.}, OV-COCO~\cite{lin2014microsoft}) but fail to generalize to vertical domain (\emph{e.g.}, ArTaxOr~\cite{drange2019arthropod}, DIOR~\cite{li2020object}, UODD~\cite{jiang2021underwater}) in the zero-shot setting.
    Although fine-tuning improves performance on vertical domain, it causes a significant degradation on general domain.
    }
    \label{fig:problem}
    \vspace{-6mm}
\end{figure}

\cwh{
Open-vocabulary object detection (OVOD)~\cite{li2022grounded,liu2024grounding,cheng2024yolo,wang2024ov} aims to enable models to detect arbitrary object categories based on the category names, including those unseen during training. Benefiting from large-scale pre-training on general object detection datasets and image-text pairs~\cite{shao2019objects365,kamath2021mdetr,plummer2015flickr30k}, existing OVOD models have achieved strong zero-shot detection performance on general scenarios (\emph{e.g.}, OV-COCO~\cite{lin2014microsoft}). 
However, when transferred to datasets with professional knowledge, their performance deteriorates sharply (see the left part of Fig.~\ref{fig:problem}), indicating that the generalization capability of existing OVOD models is still inadequate for vertical domain.
We refer to the tasks on these data as downstream tasks, which typically exhibit strong domain specificity and are designed for specialized detection scenarios where category labels carry deeper semantic granularity. For instance, instead of labeling an object simply as “butterfly”, the annotations in ArTaxOr~\cite{drange2019arthropod} may distinguish among specific butterfly species.
During large-scale pre-training, such fine-grained and domain-specific categories are often scarce and semantically limited, leading to weak textual representations and consequently misaligned vision–language semantics. This ultimately degrades detection performance when adapting OVOD models to downstream tasks.


\begin{figure*}[h]
    \centering
    \includegraphics[width=0.73\textwidth]{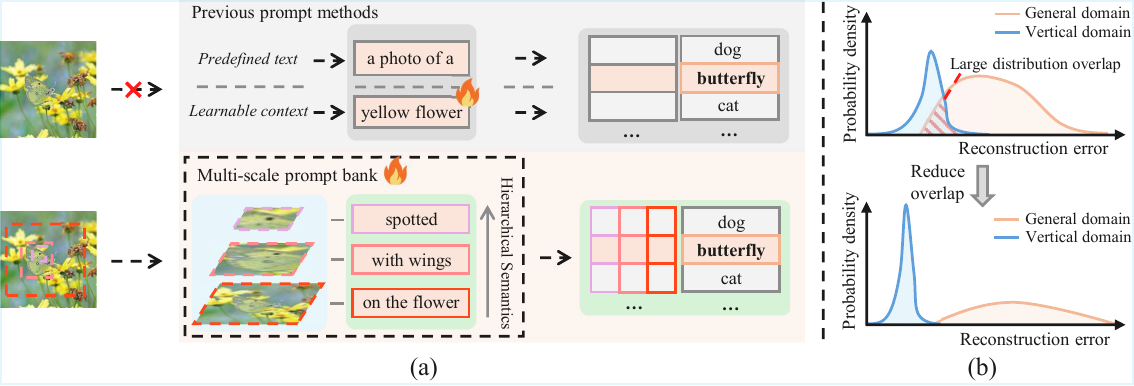} 
    \vspace{-2.5mm}
    \caption{
    Method motivation. (a) Previous methods use predefined templates or learnable vectors prepended to the category label embeddings, ignoring detailed semantics from image features. Our multi-scale prompt bank uses hierarchical semantics from multi-scale feature pyramid to select auxiliary prompts for the category labels.
    (b) The dynamic routing method~\cite{yu2024boosting} uses the reconstruction error of an input, obtained from multiple autoencoders, as an indicator of its domain. However, the large overlap in reconstruction errors across different domains can confuse the model, leading to incorrect parameter selection. Our method explicitly models both the content and the domain of the inputs, effectively reducing this overlap and enabling more accurate routing decisions.
    }
    \label{fig:motivation}
    \vspace{-5mm}
\end{figure*}

To address the issue of weak textual representations, existing methods (as illustrated in Fig.~\ref{fig:motivation} (a)) typically input a predefined prompt template (\emph{e.g.}, “a photo of a [CLS]”) into a text encoder to leverage pre-trained open-vocabulary knowledge for downstream detection, where [CLS] denotes the category label~\cite{wang2024ov}. Another common approach introduces a learnable prompt vector prepended to the category label, allowing the model to learn category-specific contextual semantics~\cite{zhou2022learning}.
However, these methods lack multi-aspect textual descriptions of visual semantics, resulting in low relevance between auxiliary prompt and visual content.
Consequently, the textual representations remain suboptimal, leading to weak visual–language alignment.
For a given image, the multi-scale feature pyramid in an OVOD model captures hierarchical semantic information, ranging from high-level contextual cues (\emph{e.g.}, flowers) to fine-grained texture details (\emph{e.g.}, spotted wings).
When this domain-specific information is used as an auxiliary prompt for the category label (\emph{e.g.}, “butterfly”), the model can enhance textual feature representations and achieve better semantic alignment. 
To this end, we propose a Multi-Scale Prompt Bank (MSPB) that serves as a bridge between image features and category labels (see Fig.~\ref{fig:motivation} (a)).
It leverages the multi-scale feature pyramid from the visual branch to select and train relevant prompts, which are concatenated with category labels and fed into the text encoder to learn domain-specific hierarchical semantics.
Training on downstream tasks in this manner enables more effective visual–language alignment across multiple semantic levels.
During inference, the model adaptively selects suitable prompts for each category label based on the input image semantics, thereby enriching textual representations and substantially improving detection performance.

Although targeted semantic augmentation on downstream tasks improves vertical domain performance, the domain gap between general domain and vertical domain causes the learned semantic augmentation to be non-generalizable.
As a result, the fine-tuned model exhibits a significant drop in mAP on general domain (see the right of Fig.~\ref{fig:problem}), causing the OVOD model to gradually lose its open-vocabulary capability.
To address this issue, we aim to enable the OVOD model to autonomously select appropriate semantic augmentation strategies according to the task characteristics and input domain.
This design allows the model to retain the semantic diversity acquired during pre-training while enhancing vertical domain understanding, thereby preserving its generalization capability across both general and downstream tasks.
Recent work, such as MoEAdapter4CL~\cite{yu2024boosting}, trains multiple autoencoders specialized for different domain distributions and uses their reconstruction errors to guide weight selection, preserving the open-vocabulary capability of the original CLIP~\cite{radford2021learning}.
However, we observe that the reconstruction errors between general domain and vertical domain highly overlap (see Fig.~\ref{fig:motivation} (b)), leading to ambiguous domain boundaries and incorrect parameter routing, which degrades OVOD performance.
Data from different domains may share highly similar visual content (\emph{e.g.}, images of cats) but differ in domain semantics (\emph{e.g.}, cartoon vs. realistic).
By focusing on content reconstruction, we can mitigate the interference caused by domain gaps in semantic representations and more accurately identify domain to select the semantic prompts correspodding to each domain. 
Motivated by this observation, we propose a Semantic-Aware Router (SAR).
It explicitly models the content and domain information of each input.
During inference, it dynamically decides whether to apply domain-specific semantic augmentation or rely on the original pre-trained representations.
This prevents parameter updates from affecting the generalization ability of the pre-trained model.

Building upon the above designs, we propose \textbf{HSA-DINO}, a parameter-efficient semantic augmentation framework with \textbf{H}ierarchical \textbf{S}emantic \textbf{A}ugmentation for enhancing open-vocabulary object detection, built upon the \textbf{DINO} architecture~\cite{zhang2022dino}.
We evaluate HSA-DINO on general domain (OV-COCO ~\cite{lin2014microsoft})
, vertical domain datasets~\cite{drange2019arthropod, li2020object, jiang2021underwater}, and modified OVOD benchmark settings. Experimental results demonstrate that our method consistently outperforms state-of-the-arts and effectively adapting pre-trained OVOD models to downstream tasks while preserving their open-vocabulary generalization capability. Our main contributions are summarized as follows:

\begin{itemize}
    \item We propose HSA-DINO, a parameter-efficient semantic augmentation framework for enhancing open-vocabulary object detection. 
    \item We design a multi-scale prompt bank that learns textual prompts from hierarchical image semantics to enhance textual representations. In addition, semantic-aware router is proposed to select appropriate semantic enhangcement prompts according to the reconstruction error of input, thereby preventing parameter updates from affecting the generalization of the pre-trained model.
    \item The experiments show that HSA-DINO outperforms state-of-the-arts, effectively adapting pre-trained OVOD models to downstream tasks while preserving open-vocabulary generalization.
\end{itemize}
}
\vspace{-2.22mm}

%% file: sec/2_related.tex
\section{Related Work}
\textbf{Open-vocabulary object detection.} 
Early methods~\cite{gu2021open,zhong2022regionclip,zang2022open} leverage CLIP~\cite{radford2021learning} to equip models with open-vocabulary capability. With the advent of large-scale image-text datasets~\cite{plummer2015flickr30k, kamath2021mdetr, sharma2018conceptual}, 
GLIP~\cite{li2022grounded} and Grounding DINO~\cite{liu2024grounding} employ pseudo labels for self-training after joint training, while YOLO-World~\cite{cheng2024yolo} constructs high-quality image-text datasets for re-joint training. OV-DINO~\cite{wang2024ov} treats the whole image as a single box associated with text descriptions. Later research reduces adaptation costs by applying parameter-efficient fine-tuning. ZiRa~\cite{deng2024zero} applies dual norm penalties to the residual detection branch of both the text and image encoders for continual learning, while MR-GDINO~\cite{dong2024mr} enhances few-shot continual learning with memory and retrieval mechanisms. However, these methods struggle to balance domian-specific adaptability and the preservation of pre-trained open-vocabulary capability.

\textbf{Prompt bank.} The prompt bank is an effective approach to enriching textual semantic representations. Unlike traditional prompt methods (\emph{e.g.}, CoOp~\cite{zhou2022learning}) that learn a fixed set of prompts, prompt bank maintains diverse learnable prompts and dynamically selects relevant ones based on input features to improve cross-modal alignment and adaptability. For example, Tip-Adapter~\cite{zhang2022tip} adopts a key-value cache for weight generation on few-shot datasets. CoCoOp~\cite{zhou2022conditional} employs a lightweight neural network for image-specific prompt adjustment. L2P~\cite{wang2022learning} extends this to continual learning with a prompt pool for task adaptation. AttriCLIP~\cite{wang2023attriclip} combines global visual features with an attribute-aware prompt library to further enhance cross-modal alignment. Despite these advances, most existing methods rely on single-scale global features for prompt selection, limiting the richness of textual representations and impairing performance on tasks like object detection that require multi-level reasoning.

\begin{figure*}[t]
\centering
\includegraphics[width=0.82\textwidth]{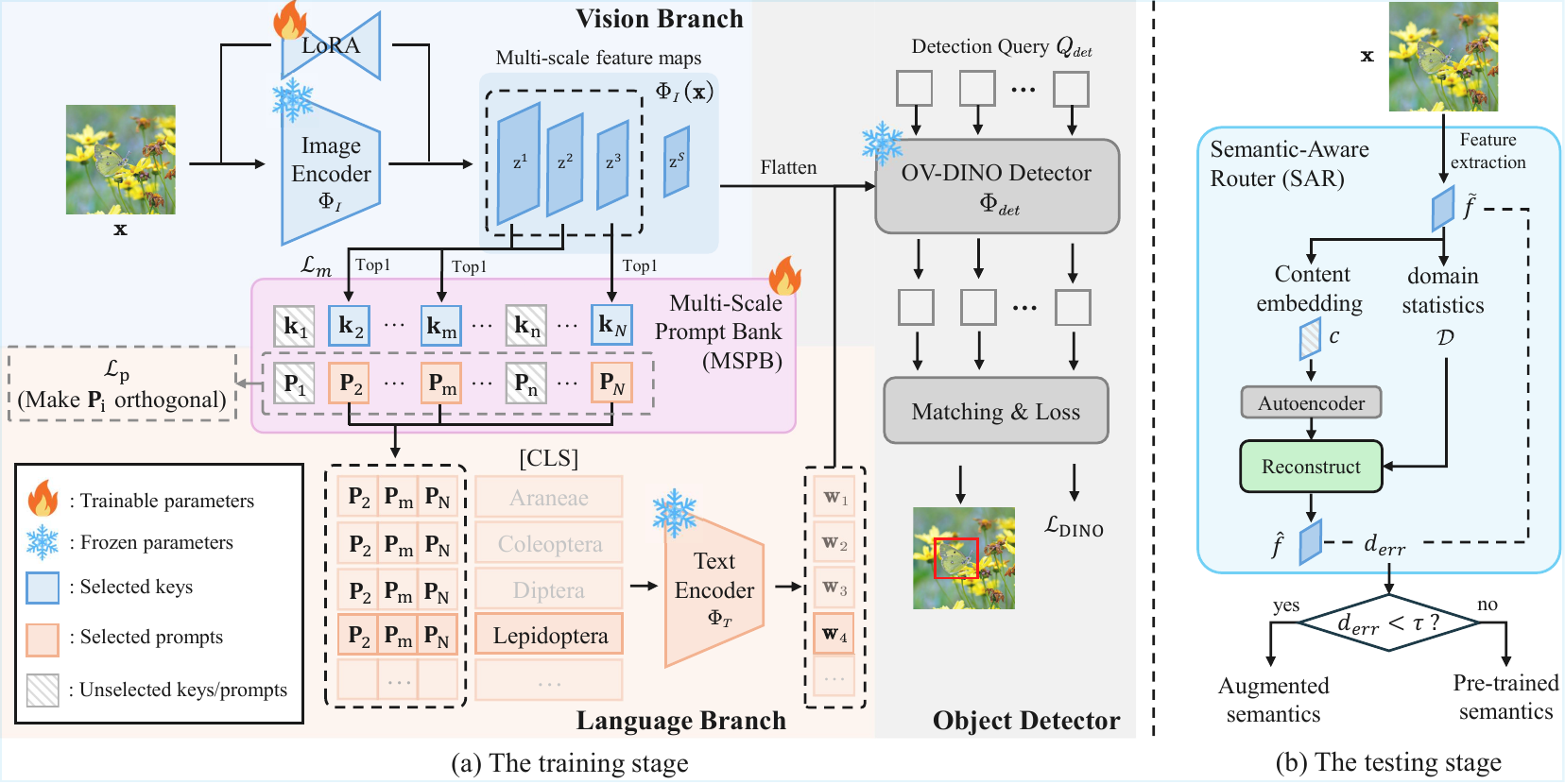} 
\vspace{-2.22mm}
\caption{Overview of the proposed HSA-DINO framework. We incorporate LoRA into the image encoder during training on downstream datasets. We also introduce a multi-scale prompt bank to further enhance the model's adaptability to downstream tasks. In the test stage, we propose a semantic-aware router that can accurately identify different data distributions, enabling more precise selection between pre-trained semantics and augmented semantics for detection.}
\label{overview}
\vspace{-3.2mm}
\end{figure*}


%% file: sec/3_method.tex
\section{Methodology}
\vspace{-1mm}
In this section, we first introduce the architecture of the proposed HSA-DINO framework, followed by detailed descriptions of its two core components: the multi-scale prompt bank and the semantic-aware router.
Finally, we present the optimization objectives of the framework.

\vspace{-1mm}
\subsection{Framework of HSA-DINO}
The overall pipeline of our proposed HSA-DINO is depicted in Fig.~\ref{overview}.
HSA-DINO is built upon OV-DINO~\cite{wang2024ov}, which consists of image encoder $\Phi_I$, text encoder $\Phi_T$ and object detector $\Phi_{det}$.
During training we keep the pre-trained parameters frozen and integrate Low-Rank Adaptation (LoRA)~\cite{hu2022lora} into the image encoder to learn hierarchical and domain-specific visual features.
For each training image $\mathbf{x}$, the LoRA-integrated image encoder $\Phi_I'$ first extracts multi-scale feature maps, which are used to select the most relevant prompts from the Multi-Scale Prompt Bank (MSPB).
The selected prompts are concatenated with the embeddings of all category labels to form the semantic-augmented text embeddings $\mathbf{t}_p$, which are then fed into the text encoder to obtain text features $E_t$.
The detector $\Phi_{det}$ receives the flattened multi-scale image features $E_i$, the text features $E_t$, and the detection queries $Q_{det}$, fusing them to produce the corresponding visual–semantic embeddings $O$ for each query, along with the predicted bounding box coordinates $B$. The classification alignment score matrix $C$ is computed by measuring the similarity between $O$ and $E_t^T$. 
The overall process of model forward is formulated as:
\begin{equation}
\setlength{\arraycolsep}{2pt}    
\begin{aligned}
E_i &= \mathrm{Flatten}(\Phi_I'(\mathbf{x})), 
& E_t &= \Phi_T(\mathbf{t}_p), \\
\{O, B\} &= \Phi_{det}(E_i, E_t, Q_{det}), 
& C &= O E_t^T.
\end{aligned}
\end{equation}
where $E_t^T$ denotes the transpose of $E_t$. 
Additionally, we train the Semantic-Aware Router (SAR) on downstream task, enabling it to accurately identify the current domain.
During testing, for each input image, the model uses the domain recognition results from SAR to determine the semantic enhancement strategy, whether to use domain-specific semantically augmented representations or to rely on the original pre-trained semantic representations, thereby ensuring that the semantic representations of OVOD remain optimal under varying domain conditions.

\vspace{-1.5mm}
\subsection{Multi-Scale Prompt Bank}
Existing methods lack multi-aspect textual descriptions of visual semantics, resulting in low relevance between auxiliary prompt and visual content.
These methods perform textual semantic augmentation on category names, but lack multi-aspect textual descriptions of visual semantics, resulting in low relevance between auxiliary prompt and visual content.
To address this limitation, we propose the Multi-Scale Prompt Bank (MSPB), which leverages multi-scale feature maps to select and concatenate prompts with category labels.
In our framework, the multi-scale feature maps are extracted from the multi-stage outputs of the Swin Transformer backbone used in OV-DINO~\cite{zhang2022dino}.
Given an input image $\mathbf{x}$, we extract the multi-scale feature maps $\{\mathbf{z}^s\}_{s=1}^S\!=\!\Phi_I'(\mathbf{x})$,
where $\mathbf{z}^s \in \mathbb{R}^{H_s \times W_s \times D}$,
$S$ denotes the number of scales, $H_s \times W_s$ is the spatial resolution at scale $s$, and $D$ is the feature dimensionality.
We then apply global average pooling and normalization to each scale to obtain
$\tilde{\mathbf{z}}^s \in \mathbb{R}^D$. To establish interactions between the semantics of images and labels, the prompt bank needs to contain both visual and textual information. Therefore, we construct the multi-scale prompt bank as $N$ (key, prompt) pairs:
\begin{equation}
    \{ \mathcal{K}, \mathcal{P} \} \stackrel{\triangle}{=} \{ (\mathbf{k}_1, \mathbf{P}_1), \ldots, (\mathbf{k}_N, \mathbf{P}_N) \},
\end{equation}
where $\{ \mathcal{K}, \mathcal{P} \}$ denotes the scale-aware prompt bank. Each $\mathbf{k}_i\!\in\!\mathbb{R}^D$ has the same dimensionality as the image feature $\tilde{\mathbf{z}}^s$, and each $\mathbf{P}_i\!=\![\mathbf{p}_i]_1\ldots[\mathbf{p}_i]_M\!\in\!\mathbb{R}^{D\times M}$ consists of $M$ learnable vectors. 
We denote the set of all keys as $\mathcal{K}\!=\!\{ \mathbf{k}_i \}_{i=1}^N$, and the set of all prompts as $\mathcal{P}\!=\!\{ \mathbf{P}_i\}_{i=1}^N$.
Here, $\mathcal{K}$ stores the scale-aware visual information, while $\mathcal{P}$ indicates the ``descriptive words'' corresponding to the keys. We aim for the feature maps at each scale to adaptively select the most relevant keys based on their scale-specific information, allowing the corresponding prompts to capture more localized and hierarchical semantics.
To this end, we compute the similarity between each scale-aware feature map $\tilde{\mathbf{z}}^s$ and each key $\mathbf{k}_i$ using cosine similarity function $\gamma$~\cite{sanh2019distilbert}. The matching keys for $\{\tilde{\mathbf{z}}^s\}_{s=1}^S
$ are selected as:
\begin{equation}
\tilde{\mathcal{K}} = \{ \mathop{\arg\max}\limits_{i\in[1,N]}\gamma(\tilde{\mathbf{z}}^s, \mathbf{k}_{i}) \}_{s=1}^S,
\end{equation}
where $\tilde{\mathcal{K}}\!\subset\!\mathcal{K}$ denotes the scale-specific selected keys. The prompts paired with these keys are then chosen as $\tilde{\mathcal{P}}\!=\!\{\mathbf{P}_{s}\}_{s=1}^S$, where $\mathbf{P}_{s}$ is the prompt corresponding to the key selected for scale $s$. These prompts are prepended to the category label embedding of $\mathbf{x}$, as illustrated in Fig.~\ref{overview}, forming the text description for the $k$-th category as:
\begin{equation}
    \mathbf{t}_p^k=\operatorname{concat}(\mathbf{P}_{1};\ldots;\mathbf{P}_{S}; [\mathrm{CLS}]_k),
\end{equation}
where $\operatorname{concat}(\cdot)$ denotes concatenation. The enhanced textual descriptions for each category are then fed into the text encoder $\Phi_T(\cdot)$ to obtain the text embeddings 
$E_t =\{\mathbf{w}_k\}_{k=1}^K =  \{\Phi_T(\mathbf{t}_p^k)\}_{k=1}^K$
, where $K$ is the number of category labels.
These text embeddings are subsequently provided to the detector $\Phi_{det}$ together with the image embeddings for final detection.

From a high-level perspective, the multi-scale prompt bank serves as a bridge between the visual and textual encoders and learns domain-specific hierarchical prompts to enrich semantic representation for enhanced detection.

\subsection{Semantic-Aware Router}
Targeted semantic augmentation on downstream tasks can improve vertical domain performance. However, the domain gap between general domain and vertical domain often makes the learned semantic augmentation non-generalizable.
To address this, we propose a Semantic-Aware Router (SAR) to dynamically determine the semantic augmentation strategy during inference, whether to apply domain-specific augmented semantic representations or to rely on the original pre-trained semantic representations.


As discussed in the introduction, domain information can interfere with distinguishing different data distributions.
To mitigate this effect, we explicitly model both the domain statistics and the content embedding, allowing the router to focus on reconstructing the content rather than being misled by domain variations.
Specifically, given an input image $\mathbf{x}$, we obtain its feature map $f=\Phi_F(\mathbf{x})$ using the feature extractor $\Phi_F(\cdot)$~\cite{he2016deep}, and apply average pooling to obtain $\tilde{f}$.
We compute the mean $\mu$ and standard deviation $\sigma$ of the feature as its domain statistics:
\begin{equation}
    \mathcal{D}\!=\!\{\mu, \sigma\},\ \ \mu\!=\!\mathrm{mean}(\tilde{f}),\ \ \sigma\!=\!\mathrm{std}(\tilde{f}).
\end{equation}
By removing the domain components from the feature, we obtain the content embedding $c$ as:
\begin{equation}
    c = \frac{\tilde{f}-\mu}{\sigma+\epsilon},
\end{equation}
which is then put into the autoencoder to obtain the reconstructed content embedding $\widehat{c}$. Finally, we reapply the domain statistics to adapt the reconstructed content back to the original domain: $\widehat{f}=\widehat{c}\cdot\sigma + \mu.$
The reconstruction error is computed as $d_{err}\!=\!|\widehat{f}\!-\!\tilde{f}|^2$, and compared against a predefined threshold $\tau$ to make the routing decision.
If $d_{err} < \tau$, the input $\mathbf{x}$ is considered to belong to the downstream distribution, and the domain-specific augmented semantic strategy is applied; otherwise, the model uses the original pre-trained semantic representations.

By explicitly modeling both domain and content information, our SAR effectively reduces the overlap in reconstruction errors across different data distributions (see Fig.~\ref{fig:motivation} (b)). This leads to more accurate routing and ensures that the semantic representations of the OVOD model 
prevent parameter updates from affecting the generalization of the pre-trained model
, thereby improving the trade-off between domain adaptation and open-vocabulary generalization.

\vspace{-1mm}
\subsection{Optimization Objectives of HSA-DINO}
Our optimization objective is improved based on OV-DINO. The optimization objectives $ \mathcal{L}_\mathrm{{DINO}}$ of OV-DINO is composed of a focal loss $\mathcal{L}_{\mathrm{cls}}$~\cite{lin2017focal}, a regression loss $\mathcal{L}_{\mathrm{box}}$~\cite{rezatofighi2019generalized}, a GIoU loss $\mathcal{L}_{\mathrm{giou}}$~\cite{rezatofighi2019generalized}, and a denoising loss $\mathcal{L}_{\mathrm{dn}}$~\cite{li2022dn}. Additionally, we propose two auxiliary losses for the training of MSPB, \emph{i.e.}, a matching loss $\mathcal{L}_{\mathrm{m}}$ and an orthogonal loss $\mathcal{L}_{\mathrm{p}}$, which are defined as:
\begin{gather}
    \mathcal{L}_{\mathrm{m}} = \sum_{s=1}^S(1-\gamma\left(\tilde{\mathbf{z}}^s, \mathbf{k}_{i_s}\right)),\\
    \mathcal{L}_\mathrm{p} = \frac{1}{N(N-1)}\sum_{n=1}^{N}\sum_{m=n+1}^{N} |\left\langle\mathbf{P}_i, \mathbf{P}_j\right\rangle|,
\end{gather}
where $\left\langle\cdot, \cdot\right\rangle$ denotes cosine similarity. $\mathcal{L}_\mathrm{m}$ encourages the selected keys to be closer to the corresponding multi-scale image features, enabling the keys to learn domain-specific knowledge from the image samples. $\mathcal{L}_{\mathrm{p}}$ orthogonalizes the embeddings of different prompts, making the learned prompts more semantically diverse. Therefore, the overall optimization objective $L$ is defined as:
\begin{equation}
\mathcal{L}\!=\!\mathcal{L}_\mathrm{DINO}\!+\!\lambda_\mathrm{m}\mathcal{L}_\mathrm{m}\!+\!\lambda_\mathrm{p}\mathcal{L}_\mathrm{p},
\end{equation}
where $\lambda_\mathrm{m}$ and $\lambda_\mathrm{p}$ control the contributions of auxiliary losses. 
For the autoencoder training in SAR,
we adopt a reconstruction loss $\mathcal{L}_\mathrm{re}$ (\emph{i.e.}, Mean Square Error).
\vspace{-1mm}

%% file: sec/4_experiments.tex
\begin{table*}[t]
\centering
\small
\setlength{\tabcolsep}{1.66mm}{
\begin{tabular}{
    c  
    c                                     
    ccc ccc ccc                           
}
\toprule[1pt]
\multirow{2}{*}{\rotatebox[origin=c]{90}{\textbf{ }}} 
& \multirow{2}{*}{\centering Method} 
& \multicolumn{3}{c}{ArTaxOr} 
& \multicolumn{3}{c}{DIOR} 
& \multicolumn{3}{c}{UODD} \\
\cmidrule(lr){3-5} \cmidrule(lr){6-8} \cmidrule(lr){9-11}
& & $\mathrm{mAP}_\mathrm{tgt}$ & $\mathrm{mAP}_\mathrm{coco}$ & $H$ & $\mathrm{mAP}_\mathrm{tgt}$ & $\mathrm{mAP}_\mathrm{coco}$ & $H$ & $\mathrm{mAP}_\mathrm{tgt}$ & $\mathrm{mAP}_\mathrm{coco}$ & $H$ \\
\midrule[0.5pt]

\multirow{4}{*}{ZS} 
& GLIP-L~\cite{li2022grounded}            & 12.6 & 51.2 & 20.2 & 4.8 & 51.2 & 8.8 & 4.6 & 51.2 & 8.4 \\
& GroundingDINO-T~\cite{liu2024grounding}   & 4.6  & 48.4 & 8.4  & 2.9 & 48.4 & 5.5 & 6.3 & 48.4 & 11.1 \\
& YOLO-World-X~\cite{cheng2024yolo}      & 2.3  & 46.6 & 4.4  & 0.1 & 46.6 & 0.2 & 0.1 & 46.6 & 0.2 \\
& OV-DINO~\cite{wang2024ov}           & 1.4  & 50.6 & 2.7  & 3.0 & 50.6 & 5.7 & 2.9 & 50.6 & 5.5 \\
\midrule[0.3pt]

\multirow{4}{*}{FFT} 
& GLIP-L~\cite{li2022grounded}                   & 73.1   & 7.6   & 13.8    & 67.5  & 15.7   & 25.5   & 36.6  & 27.8   & 31.6 \\
& GroundingDINO-T~\cite{liu2024grounding}            & 87.4 & 36.0 & 51.0 & 69.0 & 35.1 & 46.5 & 36.0 & 37.5 & 36.7 \\
& YOLO-World-X~\cite{cheng2024yolo}           & 73.6 & 0.2  & 0.4  & 68.8 & 0.1 & 0.2 & 56.1 & 0.0 & 0.0 \\
& OV-DINO~\cite{wang2024ov}                 & 85.4 & 36.1 & 50.7 & 69.4 & 37.6 & 48.8 & 53.1 & 42.2 & 47.0 \\
\midrule[0.3pt]

\multirow{4}{*}{PEFT} 
& ZiRa~\cite{deng2024zero}                              & 81.5   & 44.1   & 57.2    & 59.8  & 42.8   & 49.9   & 46.8  & 46.3   & 46.5 \\
& MR-GDINO~\cite{dong2024mr}                          & 80.7   & 13.4   & 23.0    & 62.1  & 23.7   & 34.3   & 47.4 & 0.1   & 0.2 \\
& OV-DINO~\cite{wang2024ov}      & 78.5 & 24.0 & 36.8 & 60.3 & 13.5 & 22.1 & 49.0 & 46.3 & 47.6 \\
& \textbf{HSA-DINO (ours)}                    & 76.8   & 49.9  & \textbf{60.5} & 57.3 & 49.3 & \textbf{53.0} & 48.6 & 50.6 & \textbf{49.6} \\

\bottomrule[1pt]
\end{tabular}}
\vspace{-1.88mm}
\caption{Comparison of different methods across various downstream tasks. ``ZS'' denotes the zero-shot performance of the pre-trained models, while ``FFT'' and ``PEFT'' denote the models' performance after full fine-tuning or parameter-efficient fine-tuning on downstream tasks, respectively. $\mathrm{mAP}_\mathrm{tgt}$ denotes the performance of models on corresponding target downstream datasets (\emph{i.e.}, ArTaxOr, DIOR, UODD);
$\mathrm{mAP}_\mathrm{coco}$ denotes the performance of models on the OV-COCO.
$H$ denotes the harmonic mean value defined in Eq.~\ref{eq:harmonic_mean}.
}
\label{tab:ovod-results}
\vspace{-4.0mm}
\end{table*}

\begin{table}[t]
\small
\centering
\setlength{\tabcolsep}{1.0mm}{
\begin{tabular}{
    c  
    c  
    ccc 
}
\toprule[1pt]
\multirow{2}{*}{\rotatebox[origin=c]{90}{\textbf{}}} 
& \multirow{2}{*}{\centering Method} 
& \multicolumn{3}{c}{OV-COCO$^{+}$} \\
\cmidrule(lr){3-5}
& & w ArTaxOr & w DIOR & w UODD \\
\midrule[0.5pt]

\multirow{4}{*}{ZS} 
 & GLIP-L~\cite{li2022grounded} & 48.0 & 33.2 & 49.0 \\
 & GroundingDINO-T~\cite{liu2024grounding} & 43.7 & 36.9 & 46.8 \\
 & YOLO-World-X~\cite{cheng2024yolo} & 42.9 & 36.3 & 44.9 \\
 & OV-DINO~\cite{wang2024ov} & 44.5 & 37.3 & 48.7 \\
\midrule
\multirow{4}{*}{FFT} 
 & GLIP-L~\cite{li2022grounded} & 11.7 & 13.2 & 24.5 \\
 & GroundingDINO-T~\cite{liu2024grounding} & 38.8 & 39.7 & 36.5 \\
 & YOLO-World-X~\cite{cheng2024yolo} & 5.2 & 2.0 & 1.7 \\
 & OV-DINO~\cite{wang2024ov} & 38.8 & 40.9 & 41.7 \\
\midrule
\multirow{4}{*}{PEFT} 
 & ZiRa~\cite{deng2024zero} & 46.9 & 44.4 & 46.0 \\
 & MR-GDINO~\cite{dong2024mr} & 17.1 & 24.9 & 4.9 \\
 & OVDINO~\cite{wang2024ov} & 26.2 & 20.7 & 46.2 \\
 & \textbf{HSA-DINO (ours)} & \textbf{52.3} & \textbf{50.1} & \textbf{50.5} \\
\bottomrule[1pt]
\end{tabular}
}
\caption{Comparison of mAP across different methods on OV-COCO$^{+}$.
“w” indicates the additional dataset included for joint evaluation (\emph{e.g.}, “w ArTaxOr” means OV-COCO evaluated together with ArTaxOr).}
\label{tab:ovcoco+_comparison}
\vspace{-5mm}
\end{table}




\section{Experiments}

\vspace{-1mm}
\subsection{Experimental Setup}
\textbf{Datasets.} 
We evaluate HSA-DINO on OV-COCO~\cite{lin2014microsoft} to assess its open-vocabulary generalization ability in the general domain.
OV-COCO serves as a standard benchmark in OVOD, where COCO data are excluded during pre-training, and the model is directly evaluated on the COCO dataset 
to measure its open-vocabulary capability.
we also evaluate the model on downstream tasks, including ArTaxOr, DIOR, and UODD.
These datasets cover a diverse range of vertical domains. Specifically, ArTaxOr~\cite{drange2019arthropod} is designed for arthropod detection and includes 13,993 images covering 7 categories.
DIOR~\cite{li2020object} comprises 18,245 remote sensing images across 20 categories, featuring complex scenes and diverse backgrounds.
UODD~\cite{jiang2021underwater} targets underwater object detection, containing 2,686 images of 3 marine life categories.
We additionally employ a modified OVOD benchmark, OV-COCO$^{+}$ (see the supplementary for more details), which integrates COCO with downstream datasets and expands the category labels to jointly evaluate the model’s overall generalization capability.

\textbf{Baselines.} We compare the proposed HSA-DINO with existing OVOD methods, including GLIP~\cite{li2022grounded}, Grounding DINO~\cite{liu2024grounding}, YOLO-World~\cite{cheng2024yolo} and OV-DINO~\cite{wang2024ov}, under zero-shot (ZS) and full fine-tuning (FFT) settings. FFT tunes all model parameters on downstream datasets. For parameter-efficient fine-tuning (PEFT), we compare with OV-DINO with LoRA~\cite{hu2022lora} integrated into both its image and text encoders.
We also compare ZiRa~\cite{deng2024zero} and MR-GDINO~\cite{dong2024mr} under PEFT setting.
For GLIP and YOLO-World, we use their largest model versions (\emph{i.e.}, ``-L/-X''). For the remaining methods, we use Swin-T~\cite{liu2021swin} and BERT-base~\cite{devlin2019bert} as image and text encoders, respectively.

\textbf{Implementation Details.} All downstream tasks are fine-tuned for 24 epochs with a batch size of 16. We use the AdamW~\cite{loshchilov2017decoupled} optimizer with a learning rate of 1e-3 for all parameters.
For MSPB, the prompt bank contains $N=10$ (key, prompt) pairs, each with a prompt length of $M=12$. Keys are selected using image features from the first three scales of the image encoder (\emph{i.e.}, $S=3$).
For SAR, we train the autoencoder on downstream data for 24 epochs using an SGD optimizer~\cite{robbins1951stochastic} with a learning rate of 1e-3. During testing, the routing threshold $\tau$ is fixed at 0.039 for all downstream tasks.
All baselines follow the settings of their original papers unless otherwise specified.

\textbf{Evaluation metrics.} We adopt the harmonic mean~\cite{xian2017zero} as the evaluation metric:
\begin{equation}
\label{eq:harmonic_mean}
    H = \frac{2\times\left(\mathrm{mAP}_\mathrm{tgt}\times\mathrm{mAP}_\mathrm{coco}\right)}{\mathrm{mAP}_\mathrm{tgt} + \mathrm{mAP}_\mathrm{coco}},
\end{equation}
where $\mathrm{mAP}_\mathrm{tgt}$ and $\mathrm{mAP}_\mathrm{coco}$ denote the mean Average Precision~\cite{dave2021evaluating} on the downstream task and OV-COCO, respectively. The mAP is computed as the mean AP50–95 across all categories. $H$ measures the overall generalization by balancing performance on general and vertical domains.

\begin{figure*}[t]
    \centering
    \includegraphics[width=0.95\linewidth]{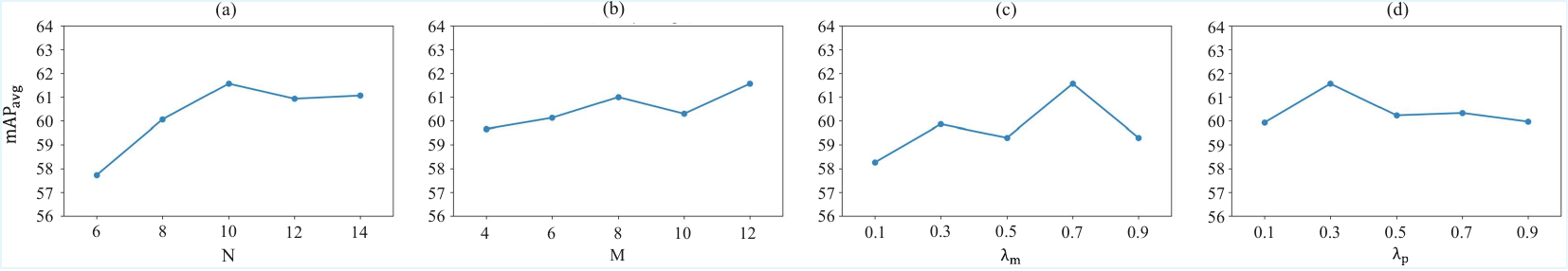}
    \vspace{-2.5mm}
    \caption{Ablation study on (a) bank size $N$, (b) prompt length $M$, (c) key matching loss weight $\lambda_{\mathrm{m}}$, and (d) orthogonal loss weight $\lambda_{\mathrm{p}}$.}
    \label{fig:ablation}
    \vspace{-4mm}
\end{figure*}

\vspace{-0.5mm}

\subsection{Main Results}
\textbf{Comparison of $H$ between OV-COCO and downstream tasks.} 
We compare our method with previous ones in Table~\ref{tab:ovod-results}. Notably, for OV-DINO under PEFT setting, we incorporate LoRA into both its image encoder and text encoder. The results show that our method outperforms previous methods in consistently achieving the highest $H$ value across all downstream tasks after fine-tuning. This indicates a superior trade-off of the model between adapting to downstream tasks and retaining open-vocabulary capability. Specifically, compared to the second-best $H$ value, our method's $H$ value increases by 3.3, 2.8, and 2.0 on the ArTaxOr, DIOR, and UODD datasets, respectively. 
We also observe that under the zero-shot (ZS) setting, all methods utilize a pre-trained OVOD model for detection, exhibiting high $\mathrm{mAP}_\mathrm{coco}$ but very low $\mathrm{mAP}_\mathrm{tgt}$. After either FFT or PEFT on downstream datasets, these methods achieve high $\mathrm{mAP}_\mathrm{tgt}$ but suffer from a significant drop in $\mathrm{mAP}_\mathrm{coco}$. For example, OV-DINO after FFT and PEFT on ArTaxOr, OV-DINO's $\mathrm{mAP}_\mathrm{coco}$ drops from 50.6 to 36.1 and 24.0, respectively. However, our method, despite being a parameter-efficient fine-tuning approach, effectively adapts the model to downstream tasks without forgetting the open-vocabulary knowledge in the pre-trained parameters. For example, comparing our method with zero-shot OV-DINO on the ArTaxOr dataset, our method effectively adapts the model to the dataset by increasing $\mathrm{mAP}_\mathrm{tgt}$ from 1.4 to 76.8, with almost no decrease in $\mathrm{mAP}_\mathrm{coco}$ (from 50.6 to 49.9). Similar trends are observed on other datasets. 

\textbf{Comparison of mAP on OV-COCO$^{+}$.} 
To further assess the scalability of our method under an expanded benchmark setting, we evaluate on OV-COCO$^{+}$. When the number of category labels is increased, categories from different tasks may interfere with each other, making joint detection more challenging.
As shown in Table~\ref{tab:ovcoco+_comparison}, HSA-DINO consistently achieves the best performance across all configurations.
In particular, it achieves $52.3$, $50.1$, and $50.5$ mAP when combined with ArTaxOr, DIOR, and UODD, respectively and surpassing the second-best method by $4.3$, $9.2$, and $1.5$ mAP. 
These results further demonstrate the superiority of HSA-DINO, achieving a superior trade-off between domain adaptability and open-vocabulary generalization.

\textbf{Comparison of different textual semantic augmentation.} We compare MSPB against three semantic augmentation strategies: (1) Predefined (“a photo of a [CLS]”)~\cite{wang2024ov}, (2) CoOp~\cite{zhou2022learning}, and (3) AttriCLIP~\cite{wang2023attriclip}.
For a fair comparison, we replace the different strategy while keeping V-LoRA (\emph{i.e.}, integrating LoRA into the image encoder) and SAR unchanged. Here, $H_\mathrm{mean}$ denotes the mean value of $H$ across all downstream datasets. 
The results in Table~\ref{tab:prompt_ablation_pro} show that the proposed MSPB achieves the highest $H$ values across all datasets after fine-tuning, further demonstrating the effectiveness of our overall method combination.

\textbf{Comparison of different routing mechanisms.} For comparison, we replace the proposed SAR with its baseline, DDAS~\cite{yu2024boosting}, which directly feeds image features into the autoencoder for reconstruction without explicitly modeling content and domain information.
As shown in Table~\ref{tab:prompt_router_ablation}, our SAR achieves higher $H$ values across all datasets.

\begin{table}[t]
\centering
\small
\setlength{\tabcolsep}{1.25mm}{
\begin{tabular}{c|ccc|c}
\toprule
Method & $H_{\text{ArTaxOr}}$ & $H_{\text{DIOR}}$ & $H_{\text{UODD}}$ & $H_{\text{mean}}$ \\
\midrule[0.5pt]
Predefined + SAR     & 54.6 & 47.3 & 47.7 & 49.9 \\
CoOp~\cite{zhou2022learning} + SAR          & 57.1 & 51.1 & 48.0 & 52.1 \\
AttriCLIP~\cite{wang2023attriclip} + SAR     & 58.8 & 51.6 & 48.5 & 53.0 \\
\textbf{MSPB + SAR (ours)} & \textbf{60.5} & \textbf{53.0} & \textbf{49.6} & \textbf{54.4} \\
\bottomrule
\end{tabular}}
\vspace{-2.4mm}
\caption{
Comparison of different textual semantic augmentation. 
}
\label{tab:prompt_ablation_pro}
\vspace{-2mm}
\end{table}

\begin{table}[t]
\centering
\small
\setlength{\tabcolsep}{1.25mm}{
\begin{tabular}{c|ccc|c}
\toprule
Method & $H_{\text{ArTaxOr}}$ & $H_{\text{DIOR}}$ & $H_{\text{UODD}}$ & $H_{\text{mean}}$ \\
\midrule[0.5pt]
MSPB + DDAS~\cite{yu2024boosting}                & 49.0      & 46.4   & 43.1   & 46.2 \\
\textbf{MSPB + SAR (ours)} & \textbf{60.5} & \textbf{53.0} & \textbf{49.6} & \textbf{54.4} \\
\bottomrule
\end{tabular}}
\vspace{-2.4mm}
\caption{
Comparison of different routing mechanisms.
}
\label{tab:prompt_router_ablation}
\vspace{-4mm}
\end{table}

\vspace{-1.5mm}
\subsection{Ablation Studies}


\textbf{Effectiveness of Different Components.}
We conduct ablations on the key components of our framework: (1) V-LoRA (\emph{i.e.}, integrating LoRA into the image encoder), (2) the proposed MSPB, and (3) the proposed SAR. As shown in Table~\ref{tab:ablation_artaxor}, introducing V-LoRA significantly improves the model’s adaptability to downstream tasks (\emph{e.g.}, $\mathrm{mAP}_\mathrm{tgt}$ increases from 1.4 to 61.6 on ArTaxOr), but still leaves considerable room for improvement. Adding MSPB further boosts downstream performance (\emph{e.g.}, from 61.6 to 79.1), demonstrating that hierarchical semantic augmentation enhances vision–language alignment.
However, without SAR, the fine-tuned weights severely degrade performance on OV-COCO due to the loss of open-vocabulary knowledge. Incorporating SAR enables dynamic semantic augmentation routing during inference, achieving a much better trade-off between $\mathrm{mAP}_\mathrm{tgt}$ and $\mathrm{mAP}_\mathrm{coco}$. Similar trends are observed across all datasets (see supplementary).

\begin{table}[t]
\centering
\small
\setlength{\tabcolsep}{1.66mm}{
\begin{tabular}{
    c c c | c c c 
}
\toprule
V-LoRA & MSPB & SAR
& $\mathrm{mAP}_\mathrm{tgt}$ & $\mathrm{mAP}_\mathrm{coco}$ & $H$ \\
\midrule[0.5pt]

           &           &           & 1.4  & 50.6 & 2.7   \\
\checkmark &           &           &  61.6     & 22.7      & 33.2  \\
           & \checkmark &           & 22.5      & 0.2      & 0.4  \\
\checkmark & \checkmark &           & \textbf{79.1}      & 0.5      & 1.0   \\
\checkmark &           & \checkmark & 59.5      & \textbf{50.4}      & 54.6 \\
           & \checkmark & \checkmark & 22      & 50.3      &  30.6  \\
\checkmark & \checkmark & \checkmark & 76.8      & 49.9      & \textbf{60.5}    \\

\bottomrule
\end{tabular}}
\vspace{-1mm}
\caption{Ablation on different components of our framework on the ArTaxOr dataset.}
\label{tab:ablation_artaxor}
\vspace{-2mm}
\end{table}

\begin{table}[t]
\centering
\small
\setlength{\tabcolsep}{1.38mm}{
\begin{tabular}{cc|ccc|c}
\toprule
$\mathcal{L}_\mathrm{m}$ & $\mathcal{L}_\mathrm{p}$ & $\mathrm{mAP}_{\text{ArTaxOr}}$ & $\mathrm{mAP}_{\text{DIOR}}$ & $\mathrm{mAP}_{\text{UODD}}$ & $\mathrm{mAP}_{\mathrm{avg}}$ \\
\midrule[0.5pt]
\checkmark &             & 78.9   & 57.4 & 48.1 & 61.5 \\
\checkmark & \checkmark  & \textbf{79.1}   & \textbf{57.7} & \textbf{48.9} & \textbf{61.9} \\
\bottomrule
\end{tabular}}
\vspace{-1mm}
\caption{Ablation on the orthogonal loss $\mathcal{L}_\mathrm{p}$.}
\label{tab:ablation_lm_lp}
\vspace{-5mm}
\end{table}

\textbf{Bank size $N$.}
We evaluate different values of the bank size $N$ in MSPB and observe that the model achieves the best $\mathrm{mAP}_{\mathrm{avg}}$ across all downstream datasets when $N=10$ (see Fig.~\ref{fig:ablation} (a)).
Here, $\mathrm{mAP}_{\mathrm{avg}}$ denotes the average performance over all downstream datasets.
When $N$ is smaller, the semantic diversity of prompts becomes limited, reducing the model’s domain adaptability.
Conversely, excessively large banks introduce redundancy among the (key, prompt) pairs, diminishing the overall performance.

\textbf{Prompt length $M$.}
We also evaluate different prompt lengths and find that the model performs best when $M\!=\!12$, as shown in Fig.~\ref{fig:ablation} (b). When the prompt length becomes smaller, the semantic capacity of the prompts is insufficient, leading to degraded performance. Due to the input length limitation of the text encoder, we do not further increase the prompt length.

\textbf{Auxiliary loss.} 
We further assess the effect of the auxiliary losses. As shown in Table~\ref{tab:ablation_lm_lp}, incorporating $\mathcal{L}_\mathrm{p}$ consistently improves the mAP of the fine-tuned model across all downstream datasets. We do not ablate $\mathcal{L}_\mathrm{m}$, since removing it would block gradient flow to MSPB.
In addition, we present the loss weight curves in Fig.~\ref{fig:ablation} (c) and (d).
From the curves, we observe that the model achieves the best performance when setting $\lambda_\mathrm{p}=0.3$ and $\lambda_\mathrm{m}=0.7$.

\begin{figure}[t]
    \centering
    \includegraphics[width=0.9\linewidth]{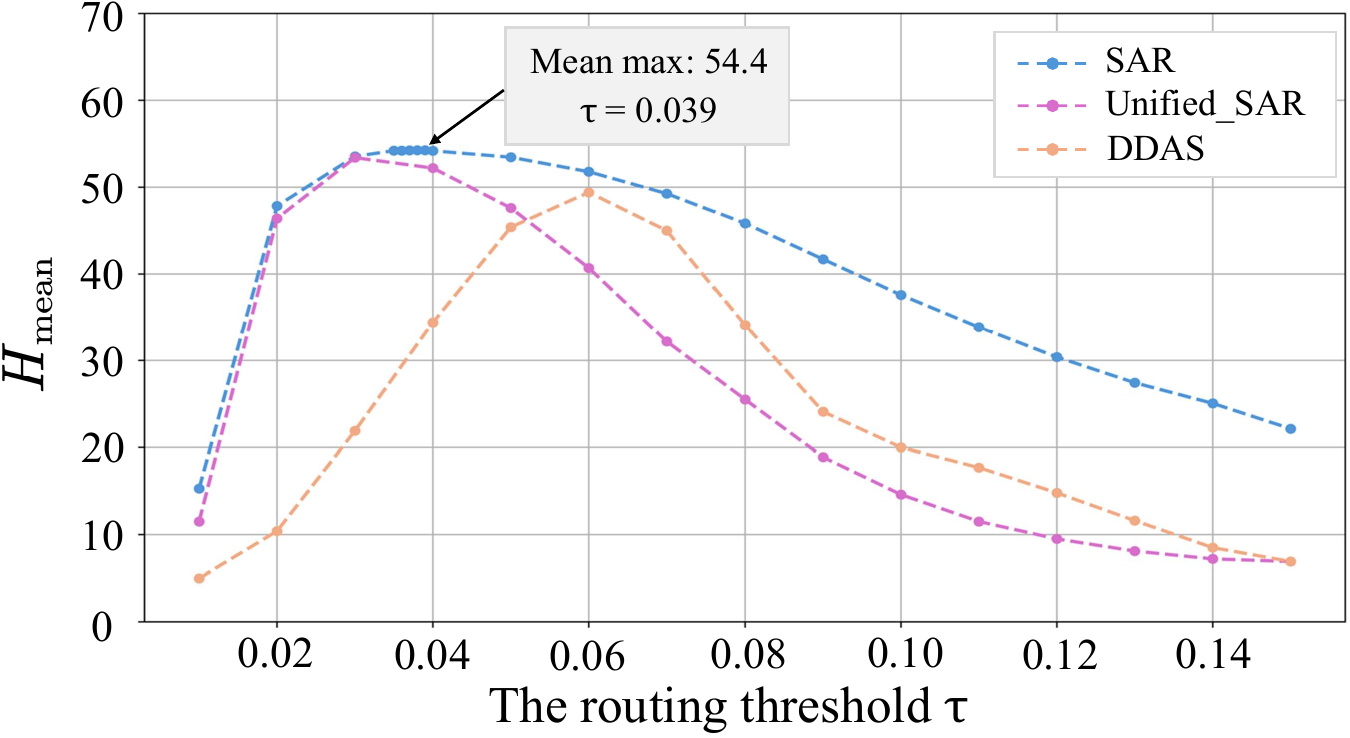}
    \vspace{-2.5mm}
    \caption{Ablation on the routing decision threshold $\tau$.}
    \label{fig:asar_threshold}
    \vspace{-2mm}
\end{figure}


\textbf{Routing threshold $\tau$.} We evaluate SAR under different values of $\tau$. 
As shown in Fig.~\ref{fig:asar_threshold}, SAR consistently outperforms DDAS across all thresholds, with the best result achieved at $\tau = 0.039$, which we use in all experiments.
To further assess generalizability, we also train a unified SAR across all downstream datasets. Although its performance is slightly lower than the task-specific variant, it still surpasses DDAS, demonstrating strong cross-domain robustness and offering a solid foundation for future extensions to more diverse domains.

\begin{figure}[t]
    \centering
    \includegraphics[width=1.0\linewidth]{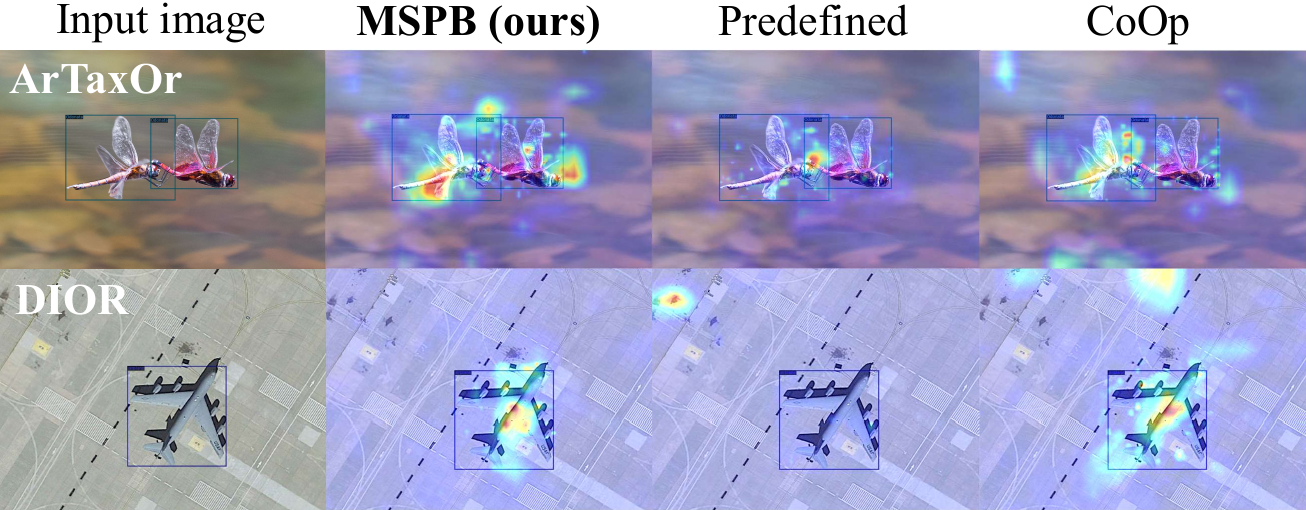}
    \vspace{-5mm}
    \caption{Visualization of the selected prompts at different textual semantic augmentation using Grad-CAM\cite{Selvaraju_2017_ICCV}.}
    \vspace{-5mm}
    \label{fig:prompt_visualization}
\end{figure}

\textbf{Visualization of prompts.} 
We visualize the prompt-wise heatmaps in Fig.~\ref{fig:prompt_visualization}.
MSPB provides richer and more diverse textual semantics, enabling prompts to align more closely with visual content.
Compared with predefined and CoOp prompts, MSPB produces more concentrated activation on target regions in single-object images (\emph{e.g.}, DIOR), and attends to a broader range of relevant areas in multi-object scenes (\emph{e.g.}, ArTaxOr).
Overall, our MSPB achieves stronger semantic grounding than existing methods that lack multi-aspect descriptions.

\begin{figure}[ht]
    \centering
    \includegraphics[width=\linewidth]{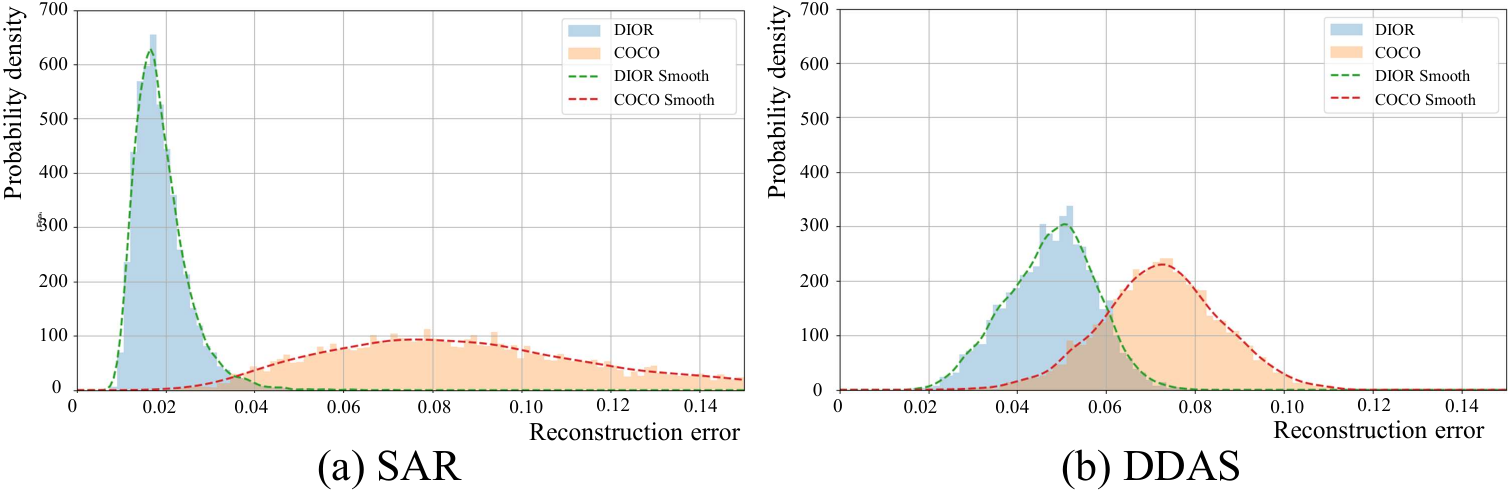}
    \vspace{-7mm}
    \caption{The probability distribution of reconstruction errors from different domain distributions (\emph{i.e.}, OV-COCO and DIOR).}
    \label{fig:asar_visual}
    \vspace{-2mm}
\end{figure}

\textbf{The reconstruction error distribution.} 
We evaluate the reconstruction errors of the routing module on test samples from COCO and DIOR. As shown in Fig.~\ref{fig:asar_visual}, our SAR produces significantly less overlap between distributions compared to DDAS. Similar trends are observed on other datasets (see supplementary). This confirms that explicitly modeling content and domain improves distribution discrimination.

\begin{figure}[t]
    \centering
    \includegraphics[width=1.0\linewidth]{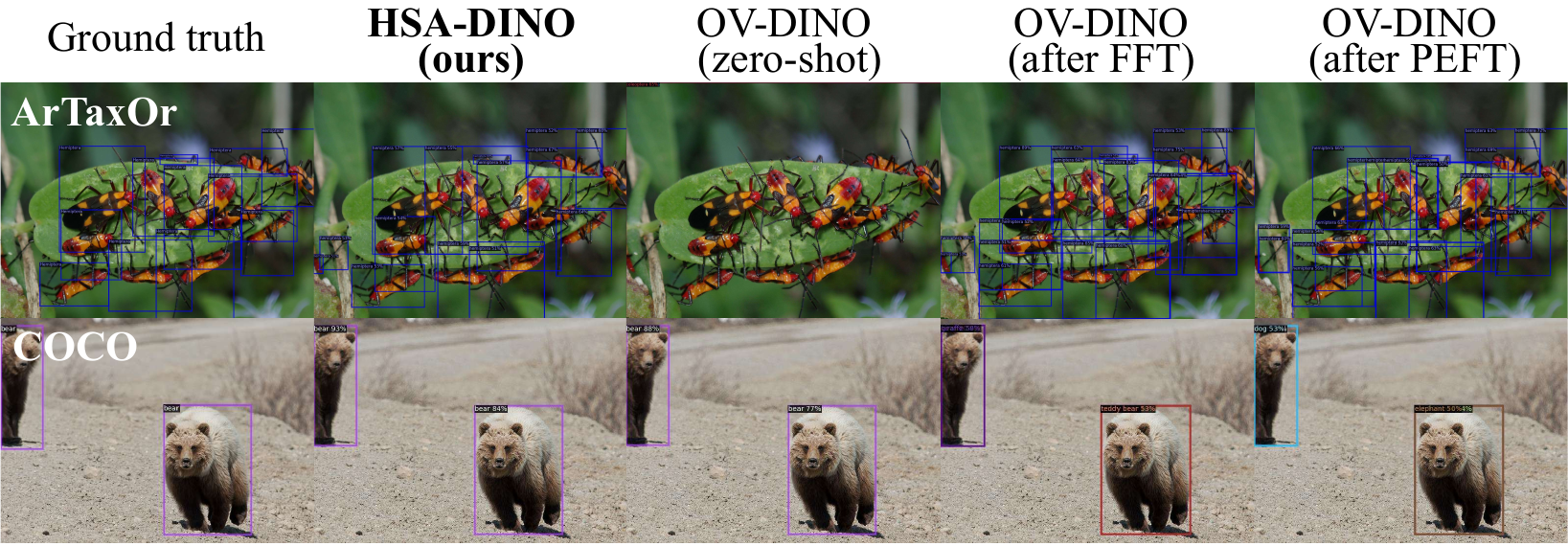}
    \vspace{-5mm}
    \caption{Visualization of detection results on selected samples from the ArTaxOr and COCO datasets using the proposed HSA-DINO and OV-DINO models at different stages.}
    \label{fig:detection_visualization}
    \vspace{-5mm}
\end{figure}

\textbf{Visualization of detection results.} As shown in Fig.~\ref{fig:detection_visualization}, the pre-trained OV-DINO (zero-shot) accurately detects objects in the COCO dataset, but fails to detect objects in the downstream dataset (\emph{i.e.}, ArTaxOr). After either FFT or PEFT, OV-DINO successfully detects objects in the downstream task, but loses its open-vocabulary capability, becoming unable to detect objects from COCO. In contrast, our proposed HSA-DINO successfully detects objects from both COCO and ArTaxOr.

\vspace{-1mm}
\section{Conclusion}
In this paper, we propose HSA-DINO, a parameter-efficient fine-tuning semantic augmentation framework for enhancing open-vocabulary object detection.
HSA-DINO effectively adapts OVOD models to downstream datasets in vertical domain while preserving the open-vocabulary capability acquired during pre-training.
By integrating a multi-scale prompt bank that leverages hierarchical semantics for textual augmentation and a semantic-aware router for adaptive semantic routing, HSA-DINO prevents parameter updates from degrading the generalization ability of the pre-trained OVOD model.
Extensive experiments demonstrate that HSA-DINO achieves an excellent balance between downstream adaptability and open-vocabulary generalization, verifying the effectiveness of our overall design.

\section{Acknowledgments}
This work was partly supported by the National Key Research and Development Program of China under Grant 2024YFE0202900; the National Natural Science Foundation of China under Grant (62436001, 62536001, 62506028); the Joint Foundation of the Ministry of Education for Innovation Team (8091B042235); the Fundamental Research Funds for the Central Universities (No. 2025JBZX064); the Postdoctoral Innovation Talent Support Program (K25M200080); and the Talent Fund of Beijing Jiaotong University (2025JBZX029 and No. 2024XKRC090).




%% file: main.bib
@String(CVPR= {Proceedings of the IEEE/CVF Conference on Computer Vision and Pattern Recognition})

@String(ICCV= {Proceedings of the IEEE/CVF international conference on computer vision})

@String(ECCV= {European conference on computer vision})

@inproceedings{zang2022open,
  title={Open-vocabulary detr with conditional matching},
  author={Zang, Yuhang and Li, Wei and Zhou, Kaiyang and Huang, Chen and Loy, Chen Change},
  booktitle=ECCV,
  pages={106--122},
  year={2022},
  organization={Springer}
}

@inproceedings{radford2021learning,
  title={Learning transferable visual models from natural language supervision},
  author={Radford, Alec and Kim, Jong Wook and Hallacy, Chris and Ramesh, Aditya and Goh, Gabriel and Agarwal, Sandhini and Sastry, Girish and Askell, Amanda and Mishkin, Pamela and Clark, Jack and others},
  booktitle={International conference on machine learning},
  pages={8748--8763},
  year={2021},
  organization={PmLR}
}

@InProceedings{Selvaraju_2017_ICCV,
author = {Selvaraju, Ramprasaath R. and Cogswell, Michael and Das, Abhishek and Vedantam, Ramakrishna and Parikh, Devi and Batra, Dhruv},
title = {Grad-CAM: Visual Explanations From Deep Networks via Gradient-Based Localization},
booktitle = ICCV,
month = {Oct},
year = {2017}
}

@inproceedings{shao2019objects365,
  title={Objects365: A large-scale, high-quality dataset for object detection},
  author={Shao, Shuai and Li, Zeming and Zhang, Tianyuan and Peng, Chao and Yu, Gang and Zhang, Xiangyu and Li, Jing and Sun, Jian},
  booktitle=ICCV,
  pages={8430--8439},
  year={2019}
}

@inproceedings{kamath2021mdetr,
  title={Mdetr-modulated detection for end-to-end multi-modal understanding},
  author={Kamath, Aishwarya and Singh, Mannat and LeCun, Yann and Synnaeve, Gabriel and Misra, Ishan and Carion, Nicolas},
  booktitle=ICCV,
  pages={1780--1790},
  year={2021}
}

@inproceedings{plummer2015flickr30k,
  title={Flickr30k entities: Collecting region-to-phrase correspondences for richer image-to-sentence models},
  author={Plummer, Bryan A and Wang, Liwei and Cervantes, Chris M and Caicedo, Juan C and Hockenmaier, Julia and Lazebnik, Svetlana},
  booktitle=ICCV,
  pages={2641--2649},
  year={2015}
}

@inproceedings{lin2014microsoft,
  title={Microsoft coco: Common objects in context},
  author={Lin, Tsung-Yi and Maire, Michael and Belongie, Serge and Hays, James and Perona, Pietro and Ramanan, Deva and Doll{\'a}r, Piotr and Zitnick, C Lawrence},
  booktitle=ECCV,
  pages={740--755},
  year={2014},
  organization={Springer}
}

@misc{drange2019arthropod,
  title={Arthropod taxonomy orders object detection dataset},
  author={Drange, Geir},
  year={2019}
}

@article{li2020object,
  title={Object detection in optical remote sensing images: A survey and a new benchmark},
  author={Li, Ke and Wan, Gang and Cheng, Gong and Meng, Liqiu and Han, Junwei},
  journal={ISPRS journal of photogrammetry and remote sensing},
  volume={159},
  pages={296--307},
  year={2020},
  publisher={Elsevier}
}

@inproceedings{jiang2021underwater,
  title={Underwater species detection using channel sharpening attention},
  author={Jiang, Lihao and Wang, Yi and Jia, Qi and Xu, Shengwei and Liu, Yu and Fan, Xin and Li, Haojie and Liu, Risheng and Xue, Xinwei and Wang, Ruili},
  booktitle={Proceedings of the ACM International Conference on Multimedia},
  pages={4259--4267},
  year={2021}
}

@inproceedings{he2016deep,
  title={Deep residual learning for image recognition},
  author={He, Kaiming and Zhang, Xiangyu and Ren, Shaoqing and Sun, Jian},
  booktitle=CVPR,
  pages={770--778},
  year={2016}
}

@article{zhang2022dino,
  title={Dino: Detr with improved denoising anchor boxes for end-to-end object detection},
  author={Zhang, Hao and Li, Feng and Liu, Shilong and Zhang, Lei and Su, Hang and Zhu, Jun and Ni, Lionel M and Shum, Heung-Yeung},
  journal={arXiv preprint},
  year={2022}
}

@article{hu2022lora,
  title={Lora: Low-rank adaptation of large language models.},
  author={Hu, Edward J and Shen, Yelong and Wallis, Phillip and Allen-Zhu, Zeyuan and Li, Yuanzhi and Wang, Shean and Wang, Lu and Chen, Weizhu and others},
  journal={International Conference on Learning Representations},
  volume={1},
  number={2},
  pages={3},
  year={2022}
}

@article{zhou2022learning,
  title={Learning to prompt for vision-language models},
  author={Zhou, Kaiyang and Yang, Jingkang and Loy, Chen Change and Liu, Ziwei},
  journal={International Journal of Computer Vision},
  volume={130},
  number={9},
  pages={2337--2348},
  year={2022},
  publisher={Springer}
}

@inproceedings{wang2023attriclip,
  title={Attriclip: A non-incremental learner for incremental knowledge learning},
  author={Wang, Runqi and Duan, Xiaoyue and Kang, Guoliang and Liu, Jianzhuang and Lin, Shaohui and Xu, Songcen and L{\"u}, Jinhu and Zhang, Baochang},
  booktitle=CVPR,
  pages={3654--3663},
  year={2023}
}

@inproceedings{yu2024boosting,
  title={Boosting continual learning of vision-language models via mixture-of-experts adapters},
  author={Yu, Jiazuo and Zhuge, Yunzhi and Zhang, Lu and Hu, Ping and Wang, Dong and Lu, Huchuan and He, You},
  booktitle=CVPR,
  pages={23219--23230},
  year={2024}
}

@inproceedings{zhong2022regionclip,
  title={Regionclip: Region-based language-image pretraining},
  author={Zhong, Yiwu and Yang, Jianwei and Zhang, Pengchuan and Li, Chunyuan and Codella, Noel and Li, Liunian Harold and Zhou, Luowei and Dai, Xiyang and Yuan, Lu and Li, Yin and others},
  booktitle=CVPR,
  pages={16793--16803},
  year={2022}
}

@article{gu2021open,
  title={Open-vocabulary object detection via vision and language knowledge distillation},
  author={Gu, Xiuye and Lin, Tsung-Yi and Kuo, Weicheng and Cui, Yin},
  journal={arXiv preprint},
  year={2021}
}

@inproceedings{sharma2018conceptual,
  author    = {Sharma, Piyush and Ding, Nan and Goodman, Sebastian and Soricut, Radu},
  title     = {Conceptual Captions: A Cleaned, Hypernymed, Image Alt-Text Dataset for Automatic Image Captioning},
  booktitle = {Proceedings of the Annual Meeting of the Association for Computational Linguistics},
  pages     = {2556--2565},
  year      = {2018}
}

@inproceedings{li2022grounded,
  title={Grounded language-image pre-training},
  author={Li, Liunian Harold and Zhang, Pengchuan and Zhang, Haotian and Yang, Jianwei and Li, Chunyuan and Zhong, Yiwu and Wang, Lijuan and Yuan, Lu and Zhang, Lei and Hwang, Jenq-Neng and others},
  booktitle=CVPR,
  pages={10965--10975},
  year={2022}
}

@inproceedings{liu2024grounding,
  title={Grounding dino: Marrying dino with grounded pre-training for open-set object detection},
  author={Liu, Shilong and Zeng, Zhaoyang and Ren, Tianhe and Li, Feng and Zhang, Hao and Yang, Jie and Jiang, Qing and Li, Chunyuan and Yang, Jianwei and Su, Hang and others},
  booktitle=ECCV,
  pages={38--55},
  year={2024},
  organization={Springer}
}

@inproceedings{cheng2024yolo,
  title={Yolo-world: Real-time open-vocabulary object detection},
  author={Cheng, Tianheng and Song, Lin and Ge, Yixiao and Liu, Wenyu and Wang, Xinggang and Shan, Ying},
  booktitle=CVPR,
  pages={16901--16911},
  year={2024}
}

@article{wang2024ov,
  title={Ov-dino: Unified open-vocabulary detection with language-aware selective fusion},
  author={Wang, Hao and Ren, Pengzhen and Jie, Zequn and Dong, Xiao and Feng, Chengjian and Qian, Yinlong and Ma, Lin and Jiang, Dongmei and Wang, Yaowei and Lan, Xiangyuan and others},
  journal={arXiv preprint},
  year={2024}
}

@inproceedings{zhou2022conditional,
  title={Conditional prompt learning for vision-language models},
  author={Zhou, Kaiyang and Yang, Jingkang and Loy, Chen Change and Liu, Ziwei},
  booktitle=CVPR,
  pages={16816--16825},
  year={2022}
}

@inproceedings{wang2022learning,
  title={Learning to prompt for continual learning},
  author={Wang, Zifeng and Zhang, Zizhao and Lee, Chen-Yu and Zhang, Han and Sun, Ruoxi and Ren, Xiaoqi and Su, Guolong and Perot, Vincent and Dy, Jennifer and Pfister, Tomas},
  booktitle=CVPR,
  pages={139--149},
  year={2022}
}

@inproceedings{xian2017zero,
  title={Zero-shot learning-the good, the bad and the ugly},
  author={Xian, Yongqin and Schiele, Bernt and Akata, Zeynep},
  booktitle=CVPR,
  pages={4582--4591},
  year={2017}
}

@article{loshchilov2017decoupled,
  title={Decoupled weight decay regularization},
  author={Loshchilov, Ilya and Hutter, Frank},
  journal={arXiv preprint},
  year={2017}
}

@inproceedings{lin2017focal,
  title={Focal loss for dense object detection},
  author={Lin, Tsung-Yi and Goyal, Priya and Girshick, Ross and He, Kaiming and Doll{\'a}r, Piotr},
  booktitle=ICCV,
  pages={2980--2988},
  year={2017}
}

@inproceedings{rezatofighi2019generalized,
  title={Generalized intersection over union: A metric and a loss for bounding box regression},
  author={Rezatofighi, Hamid and Tsoi, Nathan and Gwak, JunYoung and Sadeghian, Amir and Reid, Ian and Savarese, Silvio},
  booktitle=CVPR,
  pages={658--666},
  year={2019}
}

@inproceedings{li2022dn,
  title={Dn-detr: Accelerate detr training by introducing query denoising},
  author={Li, Feng and Zhang, Hao and Liu, Shilong and Guo, Jian and Ni, Lionel M and Zhang, Lei},
  booktitle=CVPR,
  pages={13619--13627},
  year={2022}
}

@article{dave2021evaluating,
  title={Evaluating large-vocabulary object detectors: The devil is in the details},
  author={Dave, Achal and Doll{\'a}r, Piotr and Ramanan, Deva and Kirillov, Alexander and Girshick, Ross},
  journal={arXiv preprint},
  year={2021}
}

@article{sanh2019distilbert,
  title={DistilBERT, a distilled version of BERT: smaller, faster, cheaper and lighter},
  author={Sanh, Victor and Debut, Lysandre and Chaumond, Julien and Wolf, Thomas},
  journal={arXiv preprint},
  year={2019}
}

@article{robbins1951stochastic,
  title={A stochastic approximation method},
  author={Robbins, Herbert and Monro, Sutton},
  journal={The annals of mathematical statistics},
  pages={400--407},
  year={1951},
  publisher={JSTOR}
}

@inproceedings{devlin2019bert,
  author    = {Devlin, Jacob and Chang, Ming-Wei and Lee, Kenton and Toutanova, Kristina},
  title     = {BERT: Pre-training of Deep Bidirectional Transformers for Language Understanding},
  booktitle = {Proceedings of the North American Chapter of the Association for Computational Linguistics: Human Language Technologies},
  pages     = {4171--4186},
  year      = {2019}
}

@article{deng2024zero,
  title={Zero-shot generalizable incremental learning for vision-language object detection},
  author={Deng, Jieren and Zhang, Haojian and Ding, Kun and Hu, Jianhua and Zhang, Xingxuan and Wang, Yunkuan},
  journal={Advances in Neural Information Processing Systems},
  volume={37},
  pages={136679--136700},
  year={2024}
}

@article{dong2024mr,
  title={MR-GDINO: efficient open-world continual object detection},
  author={Dong, Bowen and Huang, Zitong and Yang, Guanglei and Zhang, Lei and Zuo, Wangmeng},
  journal={arXiv preprint},
  year={2024}
}

@inproceedings{liu2021swin,
  title={Swin transformer: Hierarchical vision transformer using shifted windows},
  author={Liu, Ze and Lin, Yutong and Cao, Yue and Hu, Han and Wei, Yixuan and Zhang, Zheng and Lin, Stephen and Guo, Baining},
  booktitle=ICCV,
  pages={10012--10022},
  year={2021}
}

@inproceedings{zhang2022tip,
  title={Tip-adapter: Training-free adaption of clip for few-shot classification},
  author={Zhang, Renrui and Zhang, Wei and Fang, Rongyao and Gao, Peng and Li, Kunchang and Dai, Jifeng and Qiao, Yu and Li, Hongsheng},
  booktitle=ECCV,
  pages={493--510},
  year={2022},
  organization={Springer}
}
